\documentclass{jov_arxiv}

\usepackage{graphicx}
\usepackage{amsmath, amssymb}
\usepackage{booktabs}
\usepackage{multirow}
\usepackage{float}
\usepackage{url}
\usepackage{adjustbox}  % for scaling tables to fit width

\usepackage{tabularx,ragged2e,booktabs}
\usepackage{array} % for >{\raggedright\arraybackslash}

\newcolumntype{L}{>{\RaggedRight\arraybackslash}X}     % flexible left-aligned
\newcolumntype{C}{>{\Centering\arraybackslash}p{0.18\mytabw}} % fixed-width numeric col
\newlength{\hitH}

\usepackage{multirow,caption}
\newlength{\mytabw}
\usepackage{subcaption}
\definecolor{myblue}{rgb}{0.3,0.05,0.9}
\definecolor{grey}{rgb}{0.4, 0.4, 0.4}
\definecolor{amagenta}{rgb}{1.0, 0, 1}
\definecolor{mygreen}{rgb}{0,0.6,0} % Define a custom green color
\newcommand\ef[1]{\textcolor{myblue}{\textbf{EF:} {#1} }}

\usepackage{etoolbox}

\title{\Large Can Modern Vision Models Understand the Difference Between an Object and a Look-alike?}

\abstract{%
Recent advances in computer vision have yielded models with strong performance on recognition benchmarks; however, significant gaps remain in comparison to human perception. One subtle ability is to judge whether an image \emph{looks like} a given object without being an instance of that object. We study whether vision-language models such as CLIP \cite{radford2021learning} capture this distinction. We curated a dataset named RoLA ( \textbf{R}eal \textbf{o}r \textbf{L}ook\textbf{A}like) of real and lookalike exemplars (e.g., toys, statues, drawings, pareidolia) across multiple categories, and first evaluate a prompt-based baseline with paired “real”/“lookalike” prompts. We then estimate a direction in CLIP’s embedding space that moves representations between real and lookalike. Applying this direction to image and text embeddings improves discrimination in cross-modal retrieval on Conceptual~12M \cite{changpinyo2021conceptual}, and also enhances captions produced by a CLIP prefix captioner.%
}

\author{}{\Large Itay Cohen}{Bar Ilan University}{}{http://}{cohenitn@biu.ac.il}
\author{}{\Large Ethan Fetaya}{Bar Ilan University}{}{http://}{ethan.fetaya@biu.ac.il}
\author{}{\Large Amir Rosenfeld}{Bar Ilan University}{}{http://}{amir.rosenfeld@biu.ac.il}

\begin{document}
\maketitle

% ---------------------- Main text ----------------------
\section{Introduction}

In recent years, significant progress in computer vision has enabled systems to achieve human-level performance in various tasks, including object detection, identification, and recognition \cite{he2016deep, dosovitskiy2020image, radford2021learning, kirillov2023segment}. Studies have demonstrated that models such as CLIP \cite{radford2021learning} and SAM \cite{kirillov2023segment} can perform on par with or even surpass human accuracy on specific benchmarks. However, despite these advancements, there are still noticeable differences between the behavior of computer vision models and human perception. For example, vision models can generalize poorly on images from domains they did not see during training compared to humans. 

\begin{figure}[!htbp]
% \begin{figure}[H]
    \centering
    % First row
    \begin{minipage}{0.23\textwidth}
        \centering
        \includegraphics[width=3cm, height=3cm]{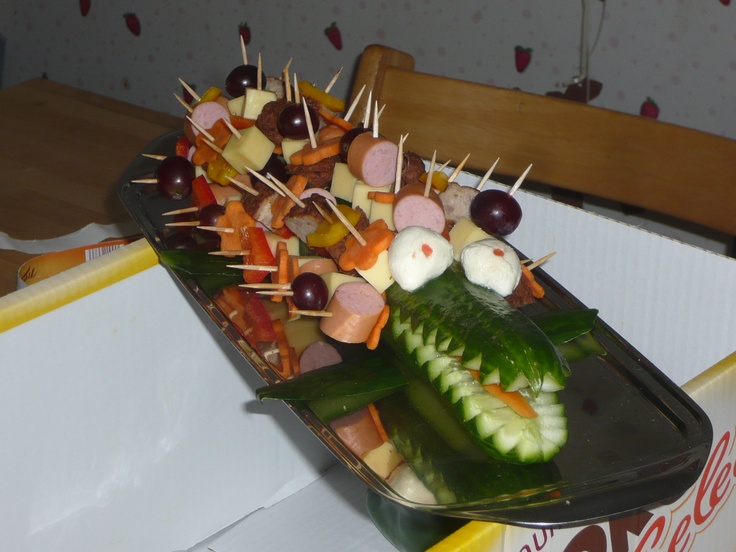}
        \label{fig:alligator_from_food}
    \end{minipage}
    \hfill
    \begin{minipage}{0.23\textwidth}
        \centering
        \includegraphics[width=3cm, height=3cm]{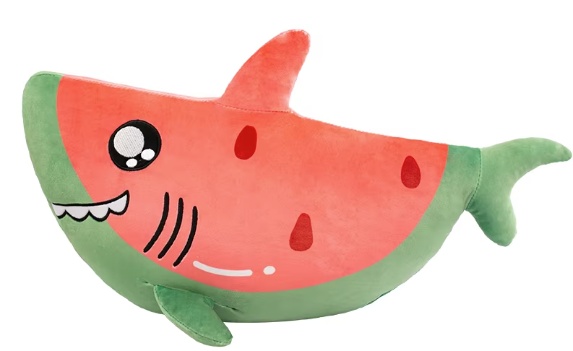}
        \label{fig:toy_shark}
    \end{minipage}
    \hfill
    \begin{minipage}{0.23\textwidth}
        \centering
        \includegraphics[width=3cm, height=3cm]{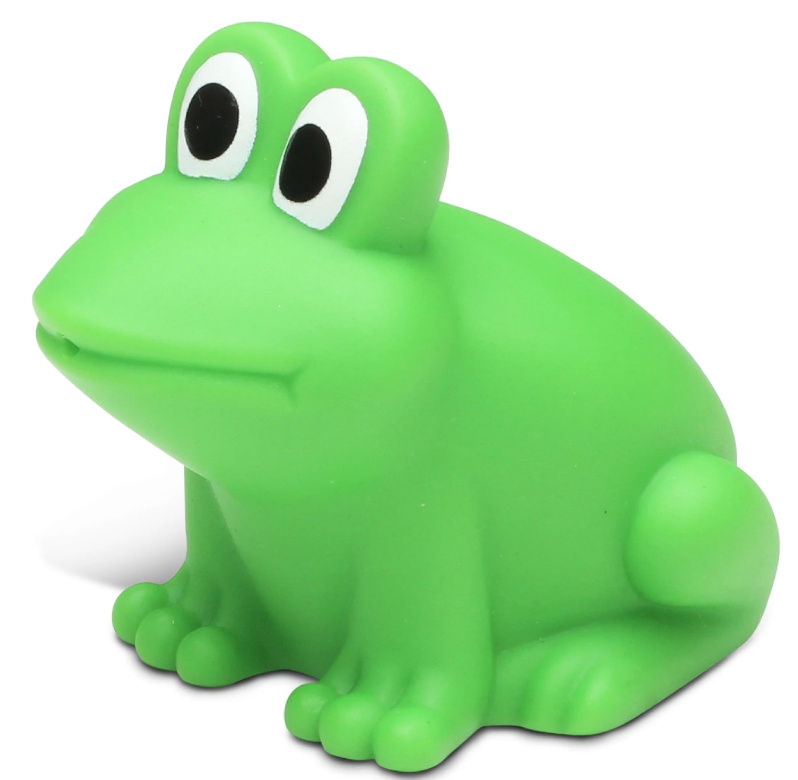}
        \label{fig:frog_toy}
    \end{minipage}
    \hfill
    \begin{minipage}{0.23\textwidth}
        \centering
        \includegraphics[width=3cm, height=3cm]{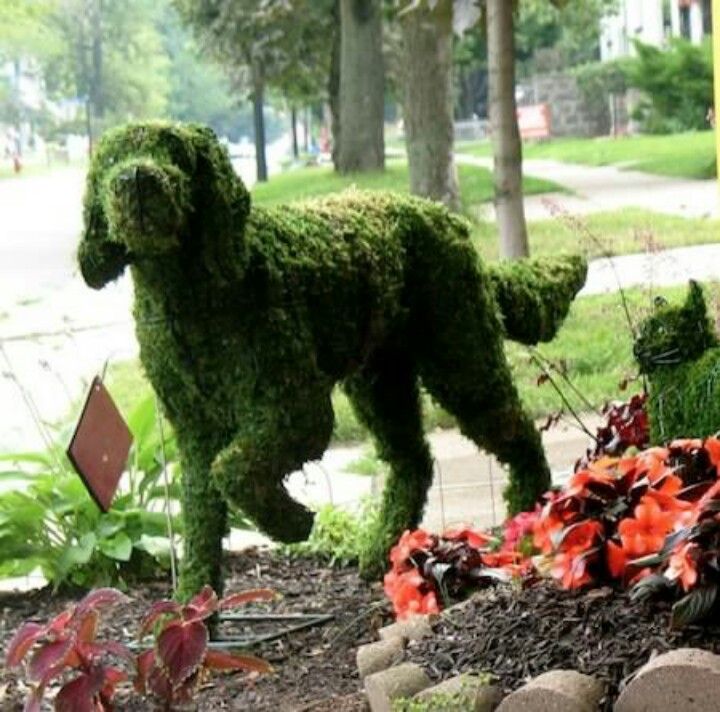}
        \label{fig:grass_statue_dog}
    \end{minipage}

    \vspace{0.17cm}
    % Second row
    \begin{minipage}{0.23\textwidth}
        \centering
        \includegraphics[width=3cm, height=3cm]{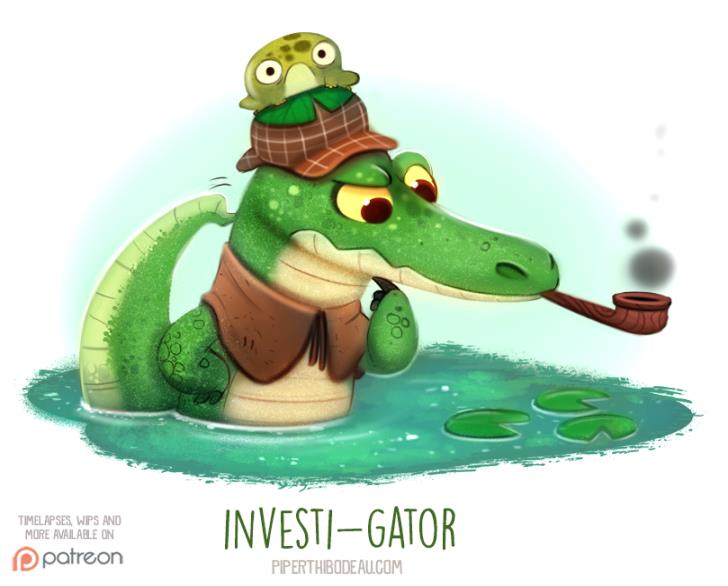}
        \label{fig:cartoon_alligator}
    \end{minipage}
    \hfill
    \begin{minipage}{0.23\textwidth}
        \centering
        \includegraphics[width=3cm, height=3cm]{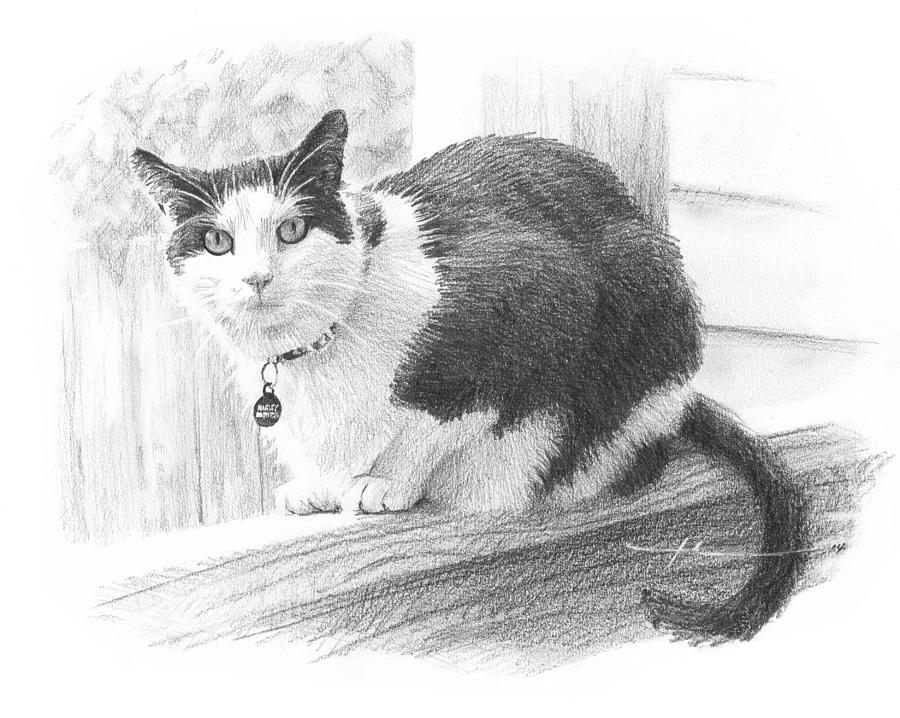}
        \label{fig:detailed_drawings_cat}
    \end{minipage}
    \hfill
    \begin{minipage}{0.23\textwidth}
        \centering
        \includegraphics[width=3cm, height=3cm]{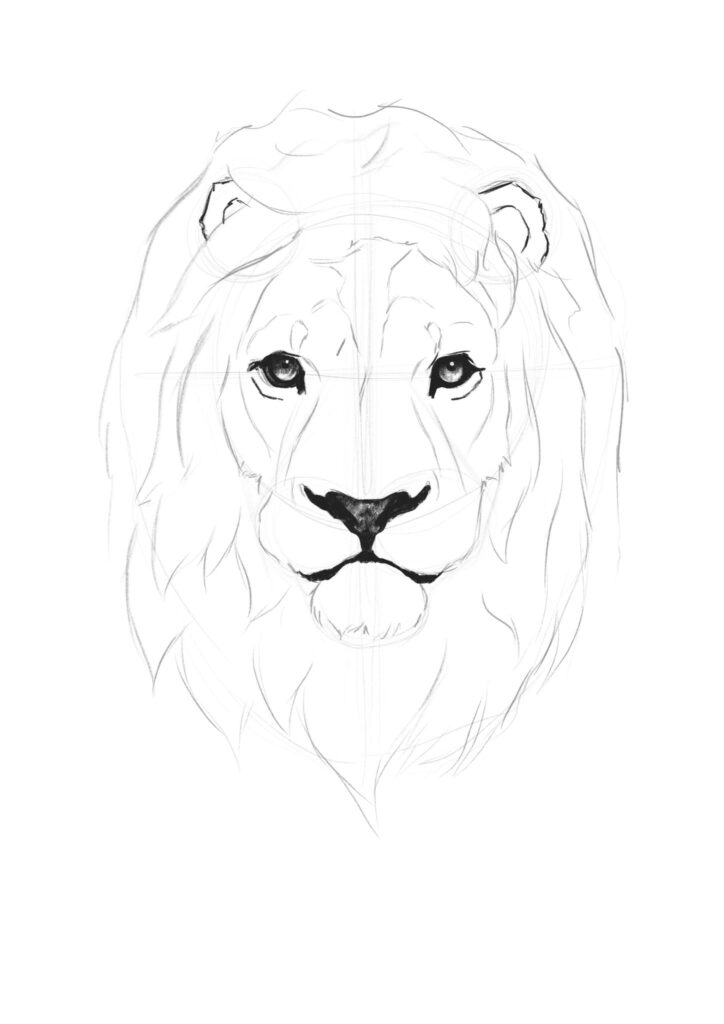}
        \label{fig:lion_drawing}
    \end{minipage}
    \hfill
    \begin{minipage}{0.23\textwidth}
        \centering
        \includegraphics[width=3cm, height=3cm]{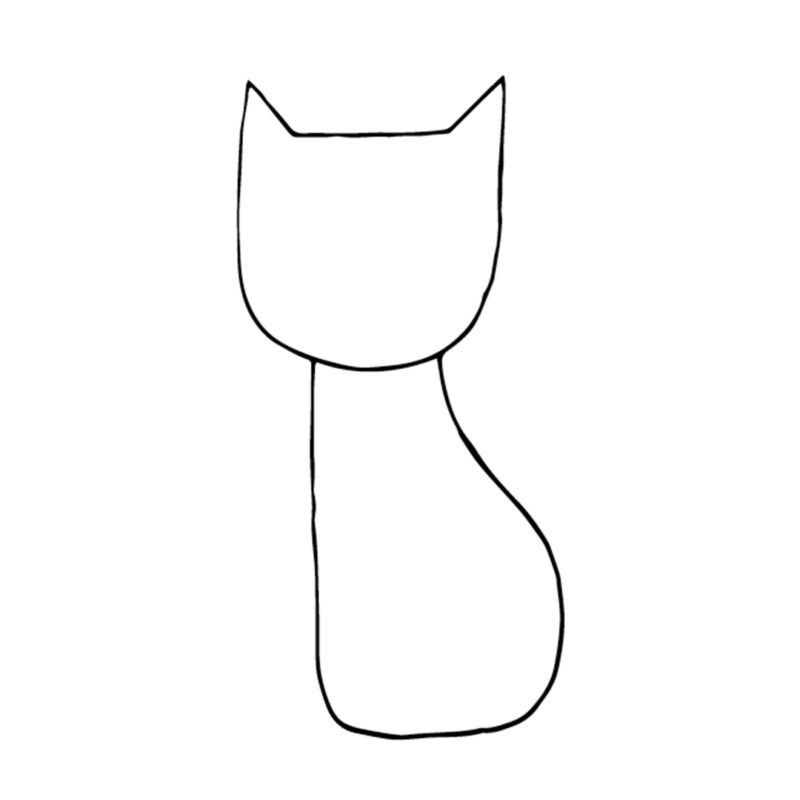}
        \label{fig:cat_scetch1}
    \end{minipage}

    \vspace{0.17cm}
    % Third row
    \begin{minipage}{0.23\textwidth}
        \centering
        \includegraphics[width=3cm, height=3cm]{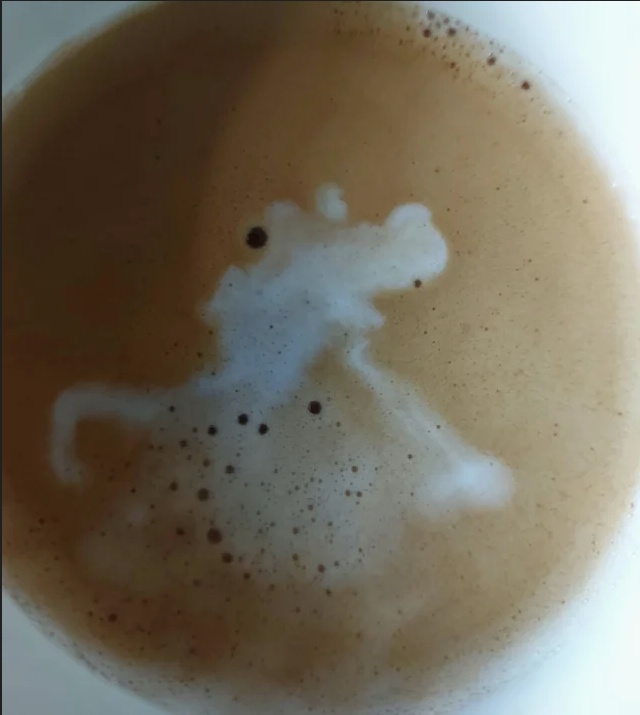}
        \label{fig:alligator_pareidolia_from_coffee}
    \end{minipage}
    \hfill
    \begin{minipage}{0.23\textwidth}
        \centering
        \includegraphics[width=3cm, height=3cm]{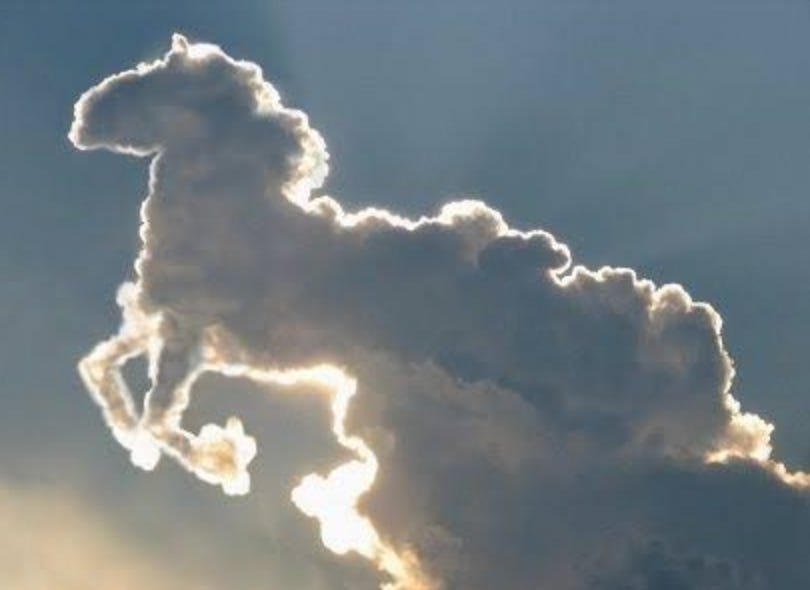}
        \label{fig:horse_pareidolia1}
    \end{minipage}
    \hfill
    \begin{minipage}{0.23\textwidth}
        \centering
        \includegraphics[width=3cm, height=3cm]{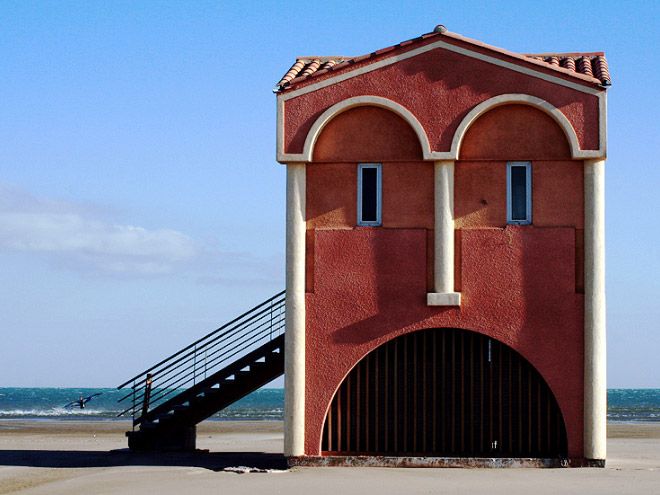}
        \label{fig:face_pareidolia_house}
    \end{minipage}
    \hfill
    \begin{minipage}{0.23\textwidth}
        \centering
        \includegraphics[width=3cm, height=3cm]{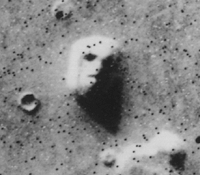}
        \label{fig:face_pareidolia_on_the_moon}
    \end{minipage}

    \caption{Lookalike examples: top row—real objects; middle—drawings; bottom—pareidolia. }
    \label{fig:look_like_grid}
\end{figure}

One subtle but critical distinction that humans naturally make is differentiating between real objects and their lookalike counterparts. Figure~\ref{fig:look_like_grid} shows the type of examples we consider: the top row presents real-world artifacts (statues and toys) that resemble target objects; the middle row presents drawings, from simple sketches to detailed illustrations; and the bottom row presents pareidolia, where random patterns evoke familiar forms. A human observer readily recognizes the object category elicited by the images, while
acknowledging that they are not real instances of those categories.
% \ar{A human observer readily recognizes both the object category elicited by the images, while acknowleding they are not real instances of those categories.}
Some lookalike images resemble multiple categories simultaneously. For instance, the second image in the top row suggests both a shark and a watermelon, yet humans still recognize that it is neither a real shark nor a real watermelon.

The purpose of this study is to test whether a vision–language model, specifically CLIP \cite{radford2021learning}, can detect this similarity while still being able to reliably distinguish real objects from lookalike counterparts. We first build a prompt-based baseline on Conceptual~12M \cite{changpinyo2021conceptual} using paired prompts for “real” and “lookalike.” We then curate a multi-category dataset, named RoLA ( \textbf{R}eal \textbf{O}r \textbf{L}ook\textbf{a}like), and estimate a real/lookalike direction in CLIP’s embedding space by taking, for each category, the difference between the mean embedding of lookalike images and the mean embedding of real images, and averaging these differences across categories. Applying this direction to image and text embeddings, we test whether discrimination improves in image retrieval on Conceptual~12M and whether image captions shift accordingly.

\paragraph{Related Work.}
Several psychophysical studies have examined face detection in the human brain, to understand how it distinguishes between real faces and face pareidolia. Using visual search tasks, the researchers found that real faces are detected the fastest, followed by face pareidolia, and the slowest for objects without faces, suggesting that the visual system contains a broadly tuned mechanism specialized for rapid face detection~\cite{keys2021visual}.
Similarly, Wardle et~al.~\cite{wardle2020rapid} demonstrated that the brain first responds to something that merely \emph{resembles} a face, but within
\(\sim\!250\,\mathrm{ms}\) determines whether it is real or just look-alike. The longer processing time for pareidolia compared to real faces indicates that different mechanisms are engaged during the identification phase: an initial, automatic face detection response followed by a slower, corrective stage that reclassifies the stimulus as non-face, highlighting that face perception involves distinct and sequential neural processes rather than a single homogeneous mechanism.
These findings motivate us to examine pareidolia from a CV (computer vision) perspective. If humans can distinguish between real and illusory faces, perhaps models can also distinguish between \emph{look-alike} and \emph{real} objects more generally.

% Building on this idea, other studies have explored pareidolia from a computer vision perspective. 
% Hamilton et~al.~\cite{hamilton2024seeing} introduced a dataset of about 5,000 face pareidolia images and examined how human face detectors exhibit face pareidolia responses, revealing a significant behavioral gap between humans and machines. 
% They found that fine-tuning models with a small number of animal face images accounts for roughly half of this gap—substantially improving pareidolic face detection compared to training on pareidolia images alone. 
% This suggests that sensitivity to animal faces may underlie the emergence of pareidolia in vision systems and that knowledge from one class can generalize to visually similar ones (e.g., using dog images to improve recognition of cats). 
% While \cite{hamilton2024seeing} focused on face pareidolia, our study extends this idea to the broader phenomenon across multiple object categories, treating face pareidolia as one instance of the general distinction between look-alike and real objects.

Building on this idea, other studies have explored pareidolia from a CV perspective.
In \cite{hamilton2024seeing} researchers created a dataset of about 5,000 face pareidolia images and examined how human face detectors exhibit face pareidolia responses, revealing a significant behavioral gap between humans and machines. 
They found that fine-tuning models with a small number of animal face images accounts for roughly half of this gap—substantially improving pareidolic face detection compared to training on pareidolia images alone.
This suggests that sensitivity to animal faces may explain the emergence of pareidolia in vision systems. This finding also inspires the idea that  knowledge from one class can help train models for visually similar classes that were not explicitly studied (e.g., using dog images to improve recognition of cats).
Although \cite{hamilton2024seeing} specifically focus on face pareidolia, our study addresses the broader phenomenon of pareidolia in diverse classes, treating it as a particular case of the general distinction between look-alike and real objects.

\section{Methods}

\noindent\textbf{Overview.}
In this paper, we curated the \textbf{RoLA} dataset and used it to identify a \emph{Real/Look-alike Direction} that captures the transformation between real and look-alike instances across categories in CLIP’s embedding space. 
We then employed this direction for classification by comparing cosine similarities between image embeddings and shifted text embeddings, evaluating variants based on the direction vector alone, prompt-based templates, and their combination. 
Finally, we applied the same direction to both \emph{image retrieval} and \emph{image captioning} tasks, demonstrating its consistent ability to shift representations toward either real or look-alike objects.

\subsection{Dataset Curation}
We assembled the RoLA dataset of about \(1{,}200\) images, about \(700\) \emph{look-alike} and \(500\) \emph{real}, manually curated from publicly available image repositories and search engines. It spans 16 object categories (cat, dog, horse, pig, rhino, elephant, alligator, snake, chameleon, spider, grasshopper, scorpion, bike, car, house, airplane). For each category, we collected paired sets of real instances and lookalike instances. Lookalikes include toys, sculptures, and drawings, as well as images that evoke the object through pareidolia.

To reduce spurious cues, we varied material, color, texture, viewpoint, and lighting. For example, the alligator class includes statues in multiple colors and finishes so that a common representation is not driven by color alone. We also selected diverse categories to enable evaluation on held-out, previously unseen classes and to test cross-category transfer. We will release the RoLA dataset, including the metadata and recommended splits, to support reproducibility and further research.

\subsection{Prompt-Based Classification} \label{subsec:Prompt-Based Classification}
We first assess CLIP’s ability to distinguish real objects from lookalike counterparts using prompt-based classification. For each category \(c\), we define two prompt templates: a \emph{real} prompt (e.g., “A photo of a real \{object\}”) and a \emph{lookalike} prompt (e.g., “An image of an object that looks like \{object\}”). Instantiating these with the category name yields text embeddings \(t_{\text{real}}(c)\) and \(t_{\text{lookalike}}(c)\). Given an image \(x\) with CLIP image embedding \(e(x)\), we assign the label with the higher cosine similarity:

\begin{equation}
\hat{y}(x)
= \arg\max_{y \in \{\text{real},\,\text{lookalike}\}}
\frac{e(x)^{\top} t_{y}(c)}{\lVert e(x)\rVert \,\lVert t_{y}(c)\rVert}.
\label{eq:prompt-classify}
\end{equation}

We evaluated this procedure on images from the RoLA dataset. Several prompt formulations were tested for each class; the most effective pair was typically \emph{``A photo of a real \{object\}''} for real images and \emph{``An image of an object that looks like a \{object\}''} for look-alikes. Using this approach, the best accuracy achieved was 81\%, with detailed results provided in Sec.~\nameref{sec:res_class}.

\subsection{Identifying the Real/Lookalike Direction}
Prior works showed that linear directions in embedding spaces can encode semantic relations \cite{bolukbasi2016man,dunlap2024describing}. We test whether CLIP admits a direction that moves representations between real and lookalike instances. For each category \(k\), we compute the mean CLIP image embeddings of real and lookalike images,
\begin{subequations}\label{eq:class-means}
\begin{align}
\bar{\mathbf{e}}_{\text{real},k}&=\frac{1}{N_k}\sum_{i=1}^{N_k}\mathbf{e}^{(i)}_{\text{real},k}, \label{eq:mean-real}\\
\bar{\mathbf{e}}_{\text{lookalike},k}&=\frac{1}{M_k}\sum_{j=1}^{M_k}\mathbf{e}^{(j)}_{\text{lookalike},k}. \label{eq:mean-look}
\end{align}
\end{subequations}
and define the difference
\begin{equation}
\mathbf{d}_k=\bar{\mathbf{e}}_{\text{lookalike},k}-\bar{\mathbf{e}}_{\text{real},k}.
\label{eq:diff-dk}
\end{equation}
Here \(N_k\) and \(M_k\) are the numbers of real and lookalike images in category \(k\in\{1,\ldots,K\}\) indexes categories (with \(K\) total), and all embeddings lie in \(\mathbb{R}^D\). For CLIP embeddings, we use CLIP ViT-B/32 that projects both image and text features to a 512-D embedding. For estimating the class-wise means \(\bar{\mathbf{e}}_{\text{real},k}\) and \(\bar{\mathbf{e}}_{\text{lookalike},k}\), we used the training set of  \(830\) images: \(480\) look-alike and \(350\) real. \\

\paragraph{leave-one-out} To estimate the \emph{Real/Look-alike Direction}, we use the \textbf{RoLA} dataset, which contains \emph{real} and \emph{lookalike} images, in 16 classes.
To test generalization beyond the specific categories, we compute a \emph{leave-one-out} mean-difference vector: for each class \(k\) ,the direction is estimated from all other classes combined, excluding the one being tested. The leave-one-out mean-difference vector \(\hat{\mathbf{d}}^{(-k)}\):
\begin{equation}
\hat{\mathbf{d}}^{(-k)}=\frac{1}{K-1}\sum_{i\ne k}\mathbf{d}_i,
\label{eq:leave one out}
\end{equation}

\paragraph{Alternative 1} For each class \(k\), we use \emph{only} the leave-one-out mean-difference vector \(\hat{\mathbf{d}}^{(-k)}\) (Eq.~\eqref{eq:leave one out}) (no text prompt).
Given an image embedding \(\mathbf{e}\), we compute the cosine similarity \(s=\langle \mathbf e, \hat{\mathbf d}^{(-k)}\rangle\).
If \(s \ge \tau\) we predict \emph{look-alike}; otherwise \emph{real}, where \(\tau\) is a threshold.
We report overall accuracy.

\paragraph{Alternative 2.}
For each class \(c\), we compute the leave-one-out mean-difference vector \(\hat{\mathbf{d}}^{(-k)}\) (Eq.~\eqref{eq:leave one out}) and shift the prompts with a step size \(0 \leq \alpha \leq 1\) before scoring. 
As shown in Eq.~\eqref{eq:diff-dk}, moving in the positive direction makes the representation more \emph{lookalike}, whereas moving in the negative direction makes it more \emph{real}:
\begin{equation}
\tilde{P}_y(c) = (1-\alpha)\,p_y(c) + \mathrm{sign}\cdot\alpha\,\hat{\mathbf d}^{(-k)},
\label{eq:shifted prompt method}
\end{equation}
where \(\mathrm{sign}=-1\) for the shifted real prompt \(\tilde{P}_r(c)\) and \(\mathrm{sign}=1\) for the shifted look-alike prompt \(\tilde{P}_l(c)\).  
We then perform the same prompt-based classification described in Sec.~\nameref{subsec:Prompt-Based Classification}: encode the shifted prompts with CLIP, compute cosine similarity with image embeddings, and report the resulting metrics.

\paragraph{Alternative 3 (single-prompt classifiers).}
We classify by cosine similarity between one prompt and the image embedding, in two variants that differ only in how the prompt is formed:

\emph{(A) Raw prompt.} For each class \(c\), use a single prompt—either \emph{real} \(p_r(c)\) or \emph{look-alike} \(p_l(c)\). Encode with CLIP and assign the label by thresholding the cosine similarity score.

\emph{(B) Shifted prompt.} 
For each class \(c\), we shift the single prompt along the leave-one-out direction \(\hat{\mathbf d}^{(-k)}\) (Eq.~\eqref{eq:leave one out}) with a step size \(0 \leq \alpha \leq 1\), as shown in (Eq.~\eqref{eq:shifted prompt method}), where \(\mathrm{sign}=1\) for the shifted real prompt \(\tilde{P}_r(c)\) and \(\mathrm{sign}=-1\) for the shifted look-alike prompt \(\tilde{P}_l(c)\).
Then, classify as in (A).

See a summary of these alternatives in Table~\ref{tab:method-vector-summary}:

\begin{table}[H]
\centering
\footnotesize

\begin{tabularx}{\linewidth}{l X}
\toprule
\textbf{Method} & \textbf{Vector(s)} \\
\midrule

\textbf{Alt.~1: leave-one-out direction only}
& $v = \hat{\mathbf d}^{(-k)}$ \\[6pt]

\textbf{Alt.~2: shifted pair of prompts}
& Shifted prompts
\[
\hat P_\ell(c) = (1-\alpha)\,P_\ell(c) + \alpha\,\hat{\mathbf d}^{(-k)},\qquad
\hat P_r(c) = (1-\alpha)\,P_r(c) - \alpha\,\hat{\mathbf d}^{(-k)},\quad 0 \le \alpha \le 1,
\]
and effective difference vector
\[
v = \hat P_\ell(c) - \hat P_r(c)
  = (1-\alpha)\bigl(P_\ell(c) - P_r(c)\bigr) + 2\alpha\,\hat{\mathbf d}^{(-k)}.
\] \\[8pt]

\textbf{Alt.~3A: single raw prompt}
& $v = p_y(c)$ \quad (a single prompt, real or look-alike) \\[6pt]

\textbf{Alt.~3B: shifted single prompt}
& 
\[
v = (1-\alpha)\,p_y(c)
    + \mathrm{sign}\cdot\alpha\,\hat{\mathbf d}^{(-k)},\qquad 0 \le \alpha \le 1,
\]
with
\[
\mathrm{sign} =
\begin{cases}
+1, & \text{shifted real prompt},\\
-1, & \text{shifted look\mbox{-}alike prompt}.
\end{cases}
\] \\

\bottomrule
\end{tabularx}

\caption{
Summary of classifier variants in terms of an image embedding $\mathbf e$ and vector(s) $v$ in CLIP space.
All methods compute a score $s = \langle \mathbf e, v\rangle$.
For Alt.~1, we predict \emph{look-alike} if $s \ge \tau$, otherwise \emph{real}.
For Alt.~3A/B, we predict the label associated with $v$ if $s \ge \tau$, otherwise the opposite label.
For Alt.~2, the decision corresponds to a threshold $\tau = 0$ (i.e., \emph{look-alike} if $s \ge 0$, otherwise \emph{real}).
}
\label{tab:method-vector-summary}
\end{table}

%% compact option:

% \begin{table}[H]
% \centering
% \footnotesize

% \begin{tabularx}{\linewidth}{l X X}
% \toprule
% \textbf{Method} & \textbf{Vector(s)} & \textbf{Score / decision rule} \\
% \midrule

% \textbf{Alt.~1: LOO direction} 
% & $v = \hat{\mathbf d}^{(-k)}$
% & $s = \langle \mathbf e, v\rangle$ \\[4pt]

% \textbf{Alt.~2: Shifted pair of prompts} 
% & $v_r = (1-\alpha)p_r(c) - \alpha\,\hat{\mathbf d}^{(-k)}$  
%   (sign = $-1$)  
%   \newline
%   $v_l = (1-\alpha)p_l(c) + \alpha\,\hat{\mathbf d}^{(-k)}$  
%   (sign = $+1$)
% & $s_r = \langle \mathbf e, v_r\rangle$,  
%   $s_l = \langle \mathbf e, v_l\rangle$  
%   \newline
%   Predict look-alike if $s_l > s_r$, else real. \\[10pt]

% \textbf{Alt.~3A: Single raw prompt} 
% & $v = p_y(c)$
% & \multirow{2}{*}{$s = \langle \mathbf e, v\rangle$;  
% predict label $y$ if $s \ge \tau$, otherwise the opposite.} \\[4pt]

% \textbf{Alt.~3B: Shifted single prompt} 
% & $v = (1-\alpha)p_y(c) + \mathrm{sign}\cdot\alpha\,\hat{\mathbf d}^{(-k)}$  
%   \newline
%   where  
%   $\mathrm{sign}=+1$ for real,  
%   $\mathrm{sign}=-1$ for look-alike
% & \\

% \bottomrule
% \end{tabularx}

% \caption{Compact summary of classifier variants.  
% All methods compute a score $s=\langle \mathbf e, v\rangle$, where $\mathbf e$ is the CLIP embedding of the image.  
% Single-vector methods (Alt.~1, Alt.~3A/B) use thresholding; the two-prompt method (Alt.~2) compares $s_l$ vs.\ $s_r$.}
% \label{tab:method-vector-summary}
% \end{table}

%% end compact option:

\subsection{Augmented Retrieval}\label{sec:Augmented_Retrieval}

All images in the retrieval database are encoded into CLIP’s shared image-text embedding space. CLIP retrieval then converts a text query into the same space and uses its embedding to retrieve the highest-scoring images from the dataset via an approximate nearest neighbors (ANN) search. Retrieved images are ranked by cosine similarity, and performance is reported as top-\(K\) accuracy.

Given an image embedding \(\mathbf{e}\) from category \(k\), we form a transformed query
\begin{equation}
% \mathbf{e}'=\mathbf{e}\pm\alpha\,\mathbf{d}_k,
\mathbf{e}'=\mathbf{e}\pm\alpha\,\hat{\mathbf d}^{(-k)},
\label{eq:shift-image}
\end{equation}

to move between the real and lookalike sets, and then rank images within the same category by cosine similarity. Accuracy is measured by whether the nearest neighbors switch to the intended counterpart set (e.g., a real dog pushed toward the lookalike direction retrieves lookalike dogs in the top-\(K\)). To test transfer, we also use a leave-one-out direction for target category \(k\), \(\hat{\mathbf d}^{(-k)}\) (Eq.~\eqref{eq:leave one out})
and repeat the same procedure with \(\hat{\mathbf d}^{(-k)}\). The step size \(\alpha\) is selected on a validation split and then fixed for evaluation.

Given our real/lookalike direction, we ask whether it can improve text-based retrieval with CLIP.

 \paragraph{Baseline.}
For each category \(c\), we instantiate either a “real” or “look-alike” prompt and encode it with CLIP’s text encoder to obtain a text embedding \(t_y(c)\). We then retrieve the highest-scoring images from Conceptual~12M using an approximate nearest neighbors (ANN) search, rank them by cosine similarity, and report top-\(K\) accuracy.  
We further explore prompt variations to identify the formulation that yields the best retrieval results. However, even with optimized prompts, real and look-alike images are often intermixed, suggesting that CLIP’s embedding space does not fully distinguish between the two.

 \paragraph{Augmentation} We shift the query representation along the learned direction, forming
\begin{equation}
\tilde{t}_y(c)=t_y(c)\pm \alpha\,\hat{\mathbf{d}},
\label{eq:shift-text}
\end{equation}
where \(\hat{\mathbf{d}}\) is either the per–category direction \(\hat{\mathbf{d}}_k\) or the leave–one–out direction \(\hat{\mathbf{d}}^{(-k)}\) (Eq.~\eqref{eq:leave one out}). We add a positive scalar for more ``look-alike'', and a negative scalar for more real. The step size \(\alpha>0\) is chosen on a validation split and then fixed.

\subsection{Image Captioning}
We test whether the same direction also influences generated descriptions. We use a CLIP prefix captioning model \cite{Mokady2021ClipCap}, where the decoder is conditioned on the CLIP image embedding. Baseline captions are produced from the unmodified embedding \(e(x)\). For augmentation, we shift the embedding with the learned direction,
\begin{equation}
\tilde{e}(x)=e(x)\pm \alpha\,\hat{\mathbf{d}},
\label{eq:shift-caption}
\end{equation} where \(\hat{\mathbf{d}}\) is either the per–category direction \(\hat{\mathbf{d}}_k\) or the leave–one–out direction \(\hat{\mathbf{d}}^{(-k)}\) (Eq.~\eqref{eq:leave one out}), and feed \(\tilde{e}(x)\) to the captioner; decoding settings are held fixed. 
% We examine whether the shift nudges captions toward the intended target: terms like ``toy,'' ``statue,'' ``drawing,'' or ``looks like'' for lookalikes, and the absence of such qualifiers for real instances.
We then evaluate how varying the shift magnitude 
\(\alpha\) affects the semantic content of the generated captions, specifically, whether the modification systematically biases captions toward the intended target. For lookalike images, we expect increased inclusion of terms such as “toy,” “statue,” “drawing,” or “looks like,” whereas captions for real instances should omit such qualifiers.

\section{Results}

\subsection{Real vs.~Lookalike Classification}\label{sec:res_class}
For each object class, we computed the cosine similarity between image embeddings and text prompts corresponding to either the real object or its lookalike. We report two types of results. First, we compare the best-performing real prompt against all lookalike prompts (see Acc. column in Table~\ref{tab:prompt_real_vs_lookalike_best_real_and_mean_diff}). Second, we compare the best-performing lookalike prompt against all real prompts (see Acc. column in Table~\ref{tab:prompt_lookalike_fixed_with_alpha}).

\paragraph{Alternative 1 ("direction only" classifier).}
For each class \(k\) we use only the leave–one–out direction \(\hat{\mathbf d}^{(-k)}\) (Eq.~\eqref{eq:leave one out}), with no text prompt. 
Given an image embedding \(\mathbf v\), we score
\(s=\langle \mathbf e, \hat{\mathbf d}^{(-k)}\rangle\) and predict
\emph{look-alike} if \(s \ge \tau\) and \emph{real} otherwise.
We test several \(\tau\) values, the best was \(\tau=0.22\), giving overall accuracy \(87.82\%\).
These results support the hypothesis that the direction alone is a strong classifier.
However, compared to the following alternative that combines prompts with the learned shift, using both prompts and the direction yields higher accuracy. \\

\paragraph{Alternative 2} 
For each class \(c\), we compute the leave-one-out mean-difference  direction \(\hat{\mathbf d}^{(-k)}\)  (Eq.~\eqref{eq:leave one out}) and shift the prompts before scoring. We use Eq.~\eqref{eq:shifted prompt method} for the shift prompts.

We then run the same prompt-based classification as in  Sec. ~\nameref{subsec:Prompt-Based Classification}: encode the (shifted) prompts with CLIP, score the images by cosine similarity, and report the resulting metrics. every pair of "real" prompt and "look-alike" prompt, get better results when we use the Real/Lookalike Direction, as we can see in  Tables~\ref{tab:prompt_real_vs_lookalike_best_real_and_mean_diff} and ~\ref{tab:prompt_lookalike_fixed_with_alpha}.

\begin{table}[H]
\centering
\captionsetup{width=\linewidth, justification=centering}
\caption{For each (Real template, Lookalike template) pair we report the accuracy, the best $\boldsymbol{\alpha}$, and the accuracy after shifting the prompts by $\boldsymbol{\alpha}$ along the mean-difference direction.}
\label{tab:prompt_real_vs_lookalike_best_real_and_mean_diff}

\renewcommand{\arraystretch}{1.25} % slightly taller rows

\begin{adjustbox}{width=\linewidth}
\begin{tabular}{|>{\centering\arraybackslash}p{0.38\linewidth}|
                  >{\raggedright\arraybackslash}p{0.38\linewidth}|
                  c|c|c|}
\hline
\textbf{Real template} & \textbf{Lookalike template} &
\textbf{Acc.} & $\boldsymbol{\alpha}$ &
\textbf{Acc. ($\boldsymbol{\alpha}$-shift)} \\ \hline

\multirow[c]{8}{*}{A photo of a real \{\}} &
  An image of an object that looks like a \{\} & 0.81 & 0.4 & 0.89 \\ \cline{2-5}
& A photo of an object that looks like a \{\}   & 0.77 & 0.06 & 0.9 \\ \cline{2-5}
& An image of a \{\}-like object                 & 0.75 & 0.07 & 0.86 \\ \cline{2-5}
& An artificial \{\}                             & 0.74 & 0.06 & 0.88 \\ \cline{2-5}
& An image of a \{\} pareidolia                  & 0.66 & 0.09 & 0.91 \\ \cline{2-5}
& An image of a \{\} look-alike                  & 0.64 & 0.22 & 0.77 \\ \cline{2-5}
& \{\} pareidolia                                & 0.61 & 0.12 & 0.94 \\ \cline{2-5}
& A photo of a \{\} pareidolia                   & 0.59 & 0.12 & 0.92 \\ \hline
\end{tabular}
\end{adjustbox}
\end{table}

% Table 4 — fixed lookalike template; merged (centered) 2nd column
\begin{table}[H]
\centering
\captionsetup{width=\linewidth, justification=centering}
\caption{For each (Real template, Lookalike template) pair we report the accuracy, the best $\boldsymbol{\alpha}$, and the accuracy after shifting the prompts by $\boldsymbol{\alpha}$ along the mean-difference direction.}
\label{tab:prompt_lookalike_fixed_with_alpha}

\renewcommand{\arraystretch}{1.25}

\begin{adjustbox}{width=\linewidth}
\begin{tabular}{|>{\raggedright\arraybackslash}p{0.38\linewidth}|
                  >{\centering\arraybackslash}p{0.38\linewidth}|
                  c|c|c|}
\hline
\textbf{Real template} & \textbf{Lookalike template} &
\textbf{Acc.} & $\boldsymbol{\alpha}$ & \textbf{Acc. ($\boldsymbol{\alpha}$-shift)} \\ \hline

A photo of a real \{\} &
\multirow[c]{5}{*}{An image of an object that looks like a \{\}} &
0.81 & 0.04 & 0.89 \\ \cline{1-1}\cline{3-5}

A photo of a \{\} & & 
0.78 & 0.05 & 0.89 \\ \cline{1-1}\cline{3-5}

An image of a real \{\} & & 
0.73 & 0.06 & 0.9 \\ \cline{1-1}\cline{3-5}

An image of \{\} & & 
0.65 & 0.08 & 0.89 \\ \cline{1-1}\cline{3-5}

\{\} & & 
0.58 & 0.29 & 0.77 \\ \hline

\end{tabular}
\end{adjustbox}
\end{table}

% \begin{table}[H]
% \centering
% \captionsetup{width=\linewidth, justification=centering}
% \caption{Identifying the Real/Lookalike Direction: For each (Real template, Lookalike template) pair we report the Accuracy, the best $\boldsymbol{\alpha}$ (by ROC–AUC), and the accuracy after shifting the prompts by $\boldsymbol{\alpha}$ along the mean-difference direction.}
% \label{tab:prompt_lookalike_fixed_with_alpha}

% \renewcommand{\arraystretch}{1.25}

% \begin{adjustbox}{width=\linewidth}
% \begin{tabular}{|>{\raggedright\arraybackslash}p{0.38\linewidth}|
%                   >{\centering\arraybackslash}p{0.38\linewidth}|
%                   c|c|c|}
% \hline
% \textbf{Real template} & \textbf{Lookalike template} &
% \textbf{Acc.} & $\boldsymbol{\alpha}$ & \textbf{Acc. ($\boldsymbol{\alpha}$-shift)} \\ \hline

% A photo of a real \{\} &
% \multirow[c]{5}{*}{An image of an object that looks like a \{\}} &
% 0.81 & 0.2 & 0.83 \\ \cline{1-1}\cline{3-5}

% A photo of a \{\} & & 
% 0.78 & 0.2 & 0.85 \\ \cline{1-1}\cline{3-5}

% An image of a real \{\} & & 
% 0.73 & 0.4 & 0.82 \\ \cline{1-1}\cline{3-5}

% An image of \{\} & & 
% 0.65 & 0.45 & 0.83 \\ \cline{1-1}\cline{3-5}

% \{\} & & 
% 0.58 & 0.6 & 0.77 \\ \hline

% \end{tabular}
% \end{adjustbox}
% \end{table}

\begin{center}
\begin{minipage}{0.9\linewidth}
\footnotesize\centering
\emph{Acc.} = accuracy over all images.  
“($\boldsymbol{\alpha}$-shift)” indicates metrics after shifting prompts by $\alpha$ along the learned direction.
\end{minipage}
\end{center}

\paragraph{Alternative 3 (single-prompt classifiers).}
We classify using one text prompt by cosine similarity to the image embedding, in two variants that differ only in how the prompt is formed:

\emph{(A) Raw prompt.} For each class \(c\), use a single prompt—either \emph{real} \(p_r(c)\) or \emph{look-alike} \(p_l(c)\). Encode with CLIP and assign the label by thresholding the cosine similarity score. Results for \emph{real} prompts appear in Table~\ref{tab:one_prompt_from_real_prompts}. For \emph{look-alike} prompts, performance collapses across all templates/thresholds (best \(\mathrm{Acc}=0.58\)), so we omit a table and report this summary instead.

\emph{(B) Shifted prompt.} 
For each class \(c\), we shift the single prompt along the leave-one-out direction \(\hat{\mathbf d}^{(-k)}\) (Eq.~\eqref{eq:leave one out}) with a step size \(0 \leq \alpha \leq 1\), as shown in (Eq.~\eqref{eq:shifted prompt method}).
Then, classify as in (A).

% We shift each prompt along the leave-one-out direction so that the \emph{real} prompt moves toward the \emph{look-alike} domain, \(\tilde{P}_r(c)\), and the \emph{look-alike} prompt moves toward the \emph{real} domain, \(\tilde{P}_l(c)\):
% \begin{equation}
% \tilde{P}_y(c) = (1-\alpha)\,p_y(c) + \mathrm{sign}\cdot\alpha\,\hat{\mathbf{d}},
% \label{eq:Alternative-3}
% \end{equation}
% where \(0 \leq \alpha \leq 1\). 
% For \(\tilde{P}_r(c)\), \(\mathrm{sign}=1\); for \(\tilde{P}_l(c)\), \(\mathrm{sign}=-1\).
% We then classify as in (A) and evaluate accuracy using ROC–AUC.  
For real prompts, the best accuracy reached approximately \(0.95\) across all classes, while for look-alike prompts it was slightly lower (\(\sim0.94\)), with one prompt reached approximately 0.95.  
The optimal performance was obtained around \(\alpha \approx 0.75\), indicating that a stronger contribution from the direction vector \(\hat{\mathbf d}^{(-k)}\) (Eq.~\eqref{eq:leave one out}) than from the original prompt yields the best results. This is consistent with the finding that the direction vector alone discriminates between real and look-alike instances more effectively than the prompt itself. At the same time, combining the two further improves performance, showing that the prompt and the learned direction captures complementary information.

% ===================== Table 3 (3 columns, same font size) =====================
\begin{table}[H]
\centering
\captionsetup{width=\mytabw, justification=centering}
\caption{Using only one \emph{real} prompt.}
\label{tab:one_prompt_from_real_prompts}

{\small
\setlength{\tabcolsep}{6pt}\renewcommand{\arraystretch}{1.15}
\begin{tabularx}{\mytabw}{|L|C|C|}
\hline
\textbf{Real template} & \(\tau\) & \textbf{Acc.} \\ \hline
A photo of a \{\}       & 0.3 & 0.69 \\ \hline
An image of \{\}      & 0.3 & 0.64 \\ \hline
An image of a real \{\} & 0.3 & 0.65 \\ \hline
A photo of a real \{\}  & 0.3 & 0.66 \\ \hline
\{\}                    & 0.4 & 0.58 \\ \hline
”direction only” classifier  & 0.22 & 0.88 \\ \hline
\end{tabularx}
}
\end{table}

\begin{center}
\begin{minipage}{0.9\linewidth}
\footnotesize\centering
\emph{Acc.} = accuracy over all images.  
“($\boldsymbol{\alpha}$-shift)” indicates metrics after shifting prompts by $\alpha$ along the learned direction.
\end{minipage}
\end{center}

\subsection{Image Retrieval}
\begin{figure}[H]
    \centering
    \begin{minipage}{0.75\textwidth}
        \centering
        \includegraphics[width=\textwidth]{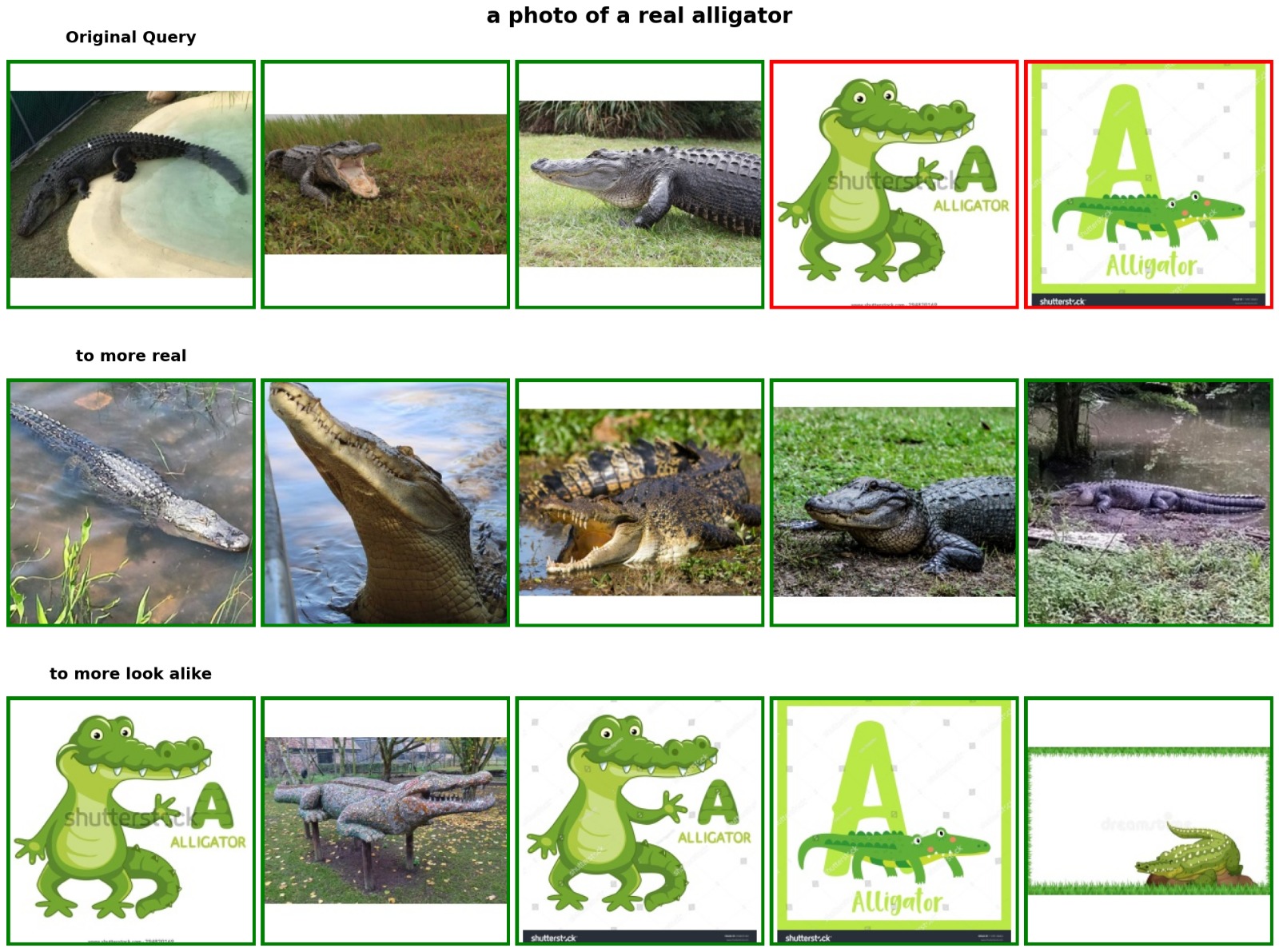}
        \caption{Retrieval with real prompt: Correct results in green, incorrect in red. Images ranked left to right by similarity; top five per row.}
        \label{fig:plus minus 1 - real alligator}
    \end{minipage}
    
    \vspace{0.6cm}
    
    \begin{minipage}{0.75\textwidth}
        \centering
        \includegraphics[width=\textwidth]{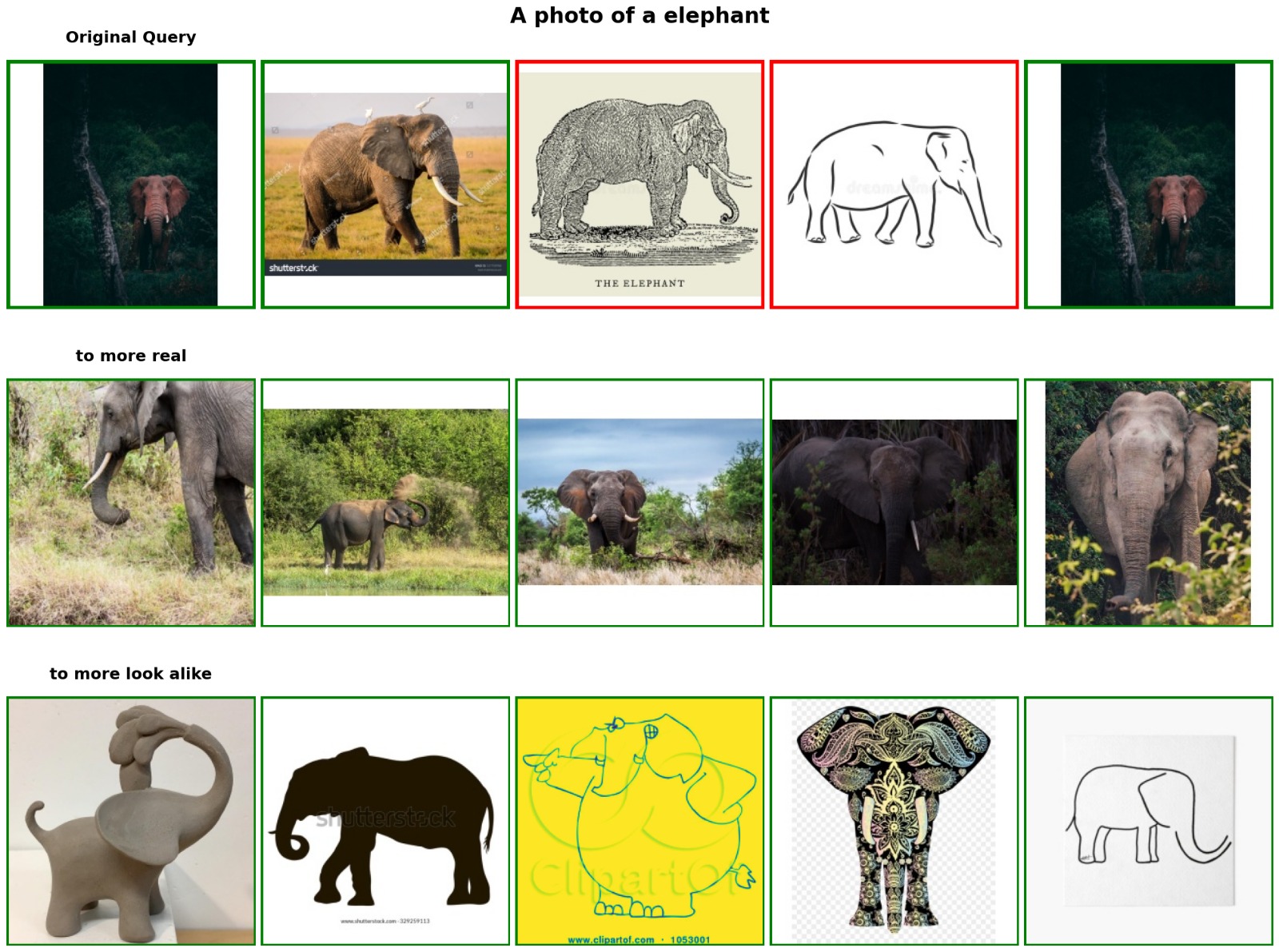}
        \caption{Retrieval with lookalike prompt: Correct results in green, incorrect in red. Images ranked left to right by similarity; top five per row.}
        \label{fig:plus minus 1 - real elephant}
    \end{minipage}
\end{figure}

In standard image retrieval, CLIP frequently makes errors such as look-alike images for real prompts (for example, drawings instead of real photo), as illustrated in the first row of Figures~\ref{fig:plus minus 1 - real alligator} and~\ref{fig:plus minus 1 - real elephant}. To mitigate these errors, we apply our proposed method, employing a leave-one-out transformation as described in Section~\nameref{sec:Augmented_Retrieval}. 

We evaluate whether this transformation enhances CLIP’s ability to discriminate between real and lookalike images in retrieval tasks. Specifically, the learned transformation is applied to the test samples and the resulting retrieval accuracy is compared against the original CLIP baseline. The improvement achieved by our method is illustrated in Figures~\ref{fig:plus minus 1 - real alligator} and~\ref{fig:plus minus 1 - real elephant}.

\paragraph{Retrieval of the top-5 images.}
For each prompt (both real and lookalike), we retrieved the five most similar images, as illustrated in the first row of Figures~\ref{fig:plus minus 1 - real alligator} and~\ref{fig:plus minus 1 - real elephant}. 
To quantitatively assess retrieval performance, we conducted a human evaluation using Amazon Mechanical Turk (AMT), see ~\nameref{sec:Human Evaluation}.
For each prompt, we computed the accuracy and corresponding 95\% binomial confidence intervals using the Wilson method, which provides bounded and statistically robust estimates for the proportion data.\\

The results for the baseline prompts and for the proposed method are summarized in Tables~\ref{tab:real_accuracy} (real prompts) and~\ref{tab:look_accuracy} (lookalike prompts). 
For the proposed method, the prompt embeddings were shifted along our learned embedding direction by $\alpha$ times the leave-one-out mean-difference vector, subtracted for real prompts (to move toward ``more real'') and added for lookalike prompts (to move toward ``more lookalike''). 
In both cases, we recomputed the accuracy and Wilson 95\% confidence intervals for the transformed embeddings.\\

As shown in Tables~\ref{tab:real_accuracy} and~\ref{tab:look_accuracy}, the proposed method consistently improves the retrieval accuracy in all types of prompts. 
For real prompts, the improvement over the baseline is substantial, while for lookalike prompts the gains vary by template, some modest, others pronounced (e.g., ``An image of a \{object\} lookalike''). 
This pattern aligns with the greater variability among the lookalike images compared to the real ones, as illustrated in Figure~\ref{fig:look_like_grid}.\\

\begin{figure}[H]
    \centering
    \begin{minipage}{0.8\textwidth}
        \centering
        \includegraphics[width=\textwidth]{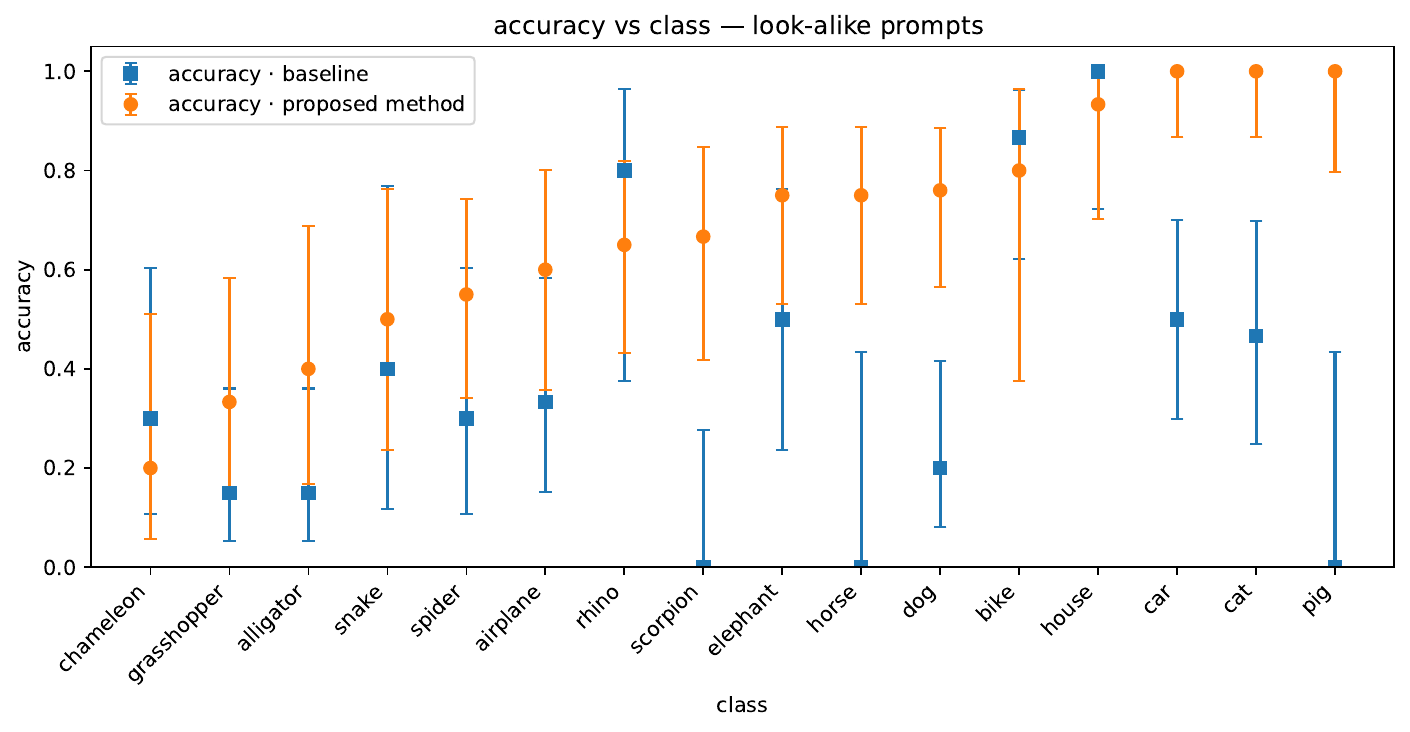}
        \caption{Retrieval with lookalike prompts: accuracy per class before/after transformation.}
        \label{fig:top5_look_alike_prompt}
    \end{minipage}
    
    \vspace{0.51cm}
    
    \begin{minipage}{0.8\textwidth}
        \centering
        \includegraphics[width=\textwidth]{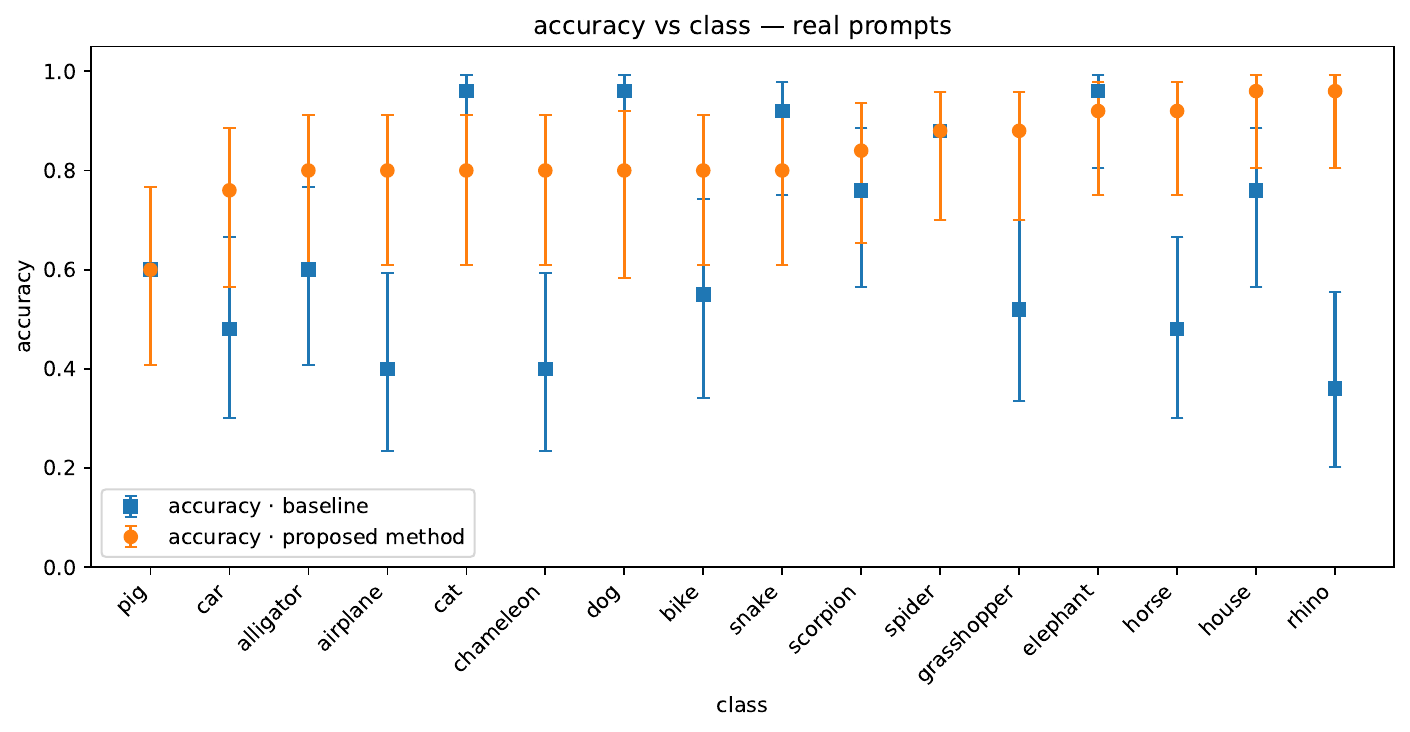}
        \caption{Retrieval with real prompts: accuracy per class before/after transformation}
        \label{fig:top5_real_prompt}
    \end{minipage}
\end{figure}

Furthermore, we present a per-class performance breakdown (Figures~\ref{fig:top5_look_alike_prompt} and~\ref{fig:top5_real_prompt}). 
Across most object classes, the proposed method consistently outperforms the baseline for both real and look-alike prompts. For real prompts, nearly all classes achieve accuracies above 80\% under the proposed method. For look-alike prompts, most classes exceed 60\% accuracy, with roughly half of the classes surpassing 70\%.

% from AMT 

\begin{table}[h]
\centering
\begin{tabular}{|l|c|c|}
\toprule
Prompt & Accuracy baseline & Accuracy proposed method \\
\midrule
a photo of a real \{object\} & 0.663 ± 0.048 & 0.833 ± 0.04 \\
A photo of a \{object\} & 0.54 ± 0.049 & 0.86 ± 0.037 \\
an image of a real \{object\} & 0.44 ± 0.049 & 0.826 ± 0.041 \\
an image of \{object\} & 0.237 ± 0.044 & 0.826 ± 0.041 \\
\{object\} & 0.231 ± 0.044 & 0.777 ± 0.044 \\
\bottomrule
\end{tabular}
\caption{Retrieval top 5 images for real prompt, for baseline and after leave-one-out transformation. Accuracy and Wilson 95\% CI.}
\label{tab:real_accuracy}
\end{table}

\begin{table}[h]
\centering
\begin{tabular}{|l|c|c|}
\toprule
Prompt & Accuracy baseline & Accuracy proposed method \\
\midrule
An image of a look-like \{object\} & 0.6 ± 0.088 & 0.67 ± 0.065 \\
An artificial \{object\} & 0.463 ± 0.058 & 0.525 ± 0.053 \\
an image of an object that looks like a \{object\} & 0.455 ± 0.064 & 0.6 ± 0.055 \\
An image of a \{object\} look-alike & 0.369 ± 0.07 & 0.721 ± 0.057 \\
a photo of an object that looks like a \{object\} & 0.361 ± 0.061 & 0.504 ± 0.053 \\
an image of a \{object\} pareidolia & 0.252 ± 0.048 & 0.509 ± 0.054 \\
\{object\} pareidolia & 0.241 ± 0.052 & 0.470 ± 0.054 \\
A photo of a \{object\} pareidolia & 0.182 ± 0.045 & 0.378 ± 0.051 \\
\bottomrule
\end{tabular}
\caption{Retrieval top 5 images for look-alike prompts, for baseline and after leave-one-out transformation.
\\Accuracy and Wilson 95\% CI.}
\label{tab:look_accuracy}
\end{table}

% from AMT 

\paragraph{Retrieval of first three changes.}
For each prompt (real and look-alike), we begin by retrieving the top five nearest images in the CC12M dataset and storing their embeddings. For each class \(c\), we then compute the leave-one-out mean-difference direction \(\hat{\mathbf d}^{(-k)}\) (Eq.~\eqref{eq:leave one out}) and iteratively shift each of these five embeddings along this direction, using the update rule in Eq.~\eqref{eq:shift-caption} with a fixed step size \(\alpha = 0.01\).

At every step, the embedding is moved slightly toward the real or look-alike direction, and we query the CC12M dataset using an approximate nearest-neighbor (ANN) index. Whenever the identity of the nearest neighbor differs from that of the previous step, we record this event as a \emph{change} and store the corresponding retrieved image. This procedure yields the sequence of the first three changes that occur as the embedding is moved along the learned direction.

% Across 500 items spanning 16 classes and multiple \(\alpha\) values, AMT judgments  (see ~\nameref{sec:Human Evaluation}) tracked our embedding transformation: small negative shifts tended to produce \emph{real} and small positive shifts \emph{lookalike}; very large shifts sometimes left the class and elicited \emph{something else}. Disagreements concentrated in visually similar categories (e.g., dog–wolf, cat–tiger; Fig.~\ref{fig:hit-disagreements-examples}). Class-specific directions worked best within their own class, while a small leave-one-out direction generalized well with only a modest drop in agreement.\\
Across 500 items spanning 16 classes and multiple \(\alpha\) values, AMT judgments  (see ~\nameref{sec:Human Evaluation}) tracked our embedding transformation: Small negative shifts tend to retrieve progressively \emph{more real} instances of the class, whereas small positive shifts reveal increasingly \emph{look-alike} counterparts. Sometimes very large shifts changed the class, which we labeled as \emph{something else}. Disagreements and unstable transitions are concentrated in visually similar categories (e.g., dog–wolf, cat–tiger;  Fig.~\ref{fig:hit-disagreements-examples}).\\

\begin{figure}[H]
    \centering

    % Row 1
    \begin{minipage}{0.9\textwidth}
        \centering
        \includegraphics[width=\textwidth]{"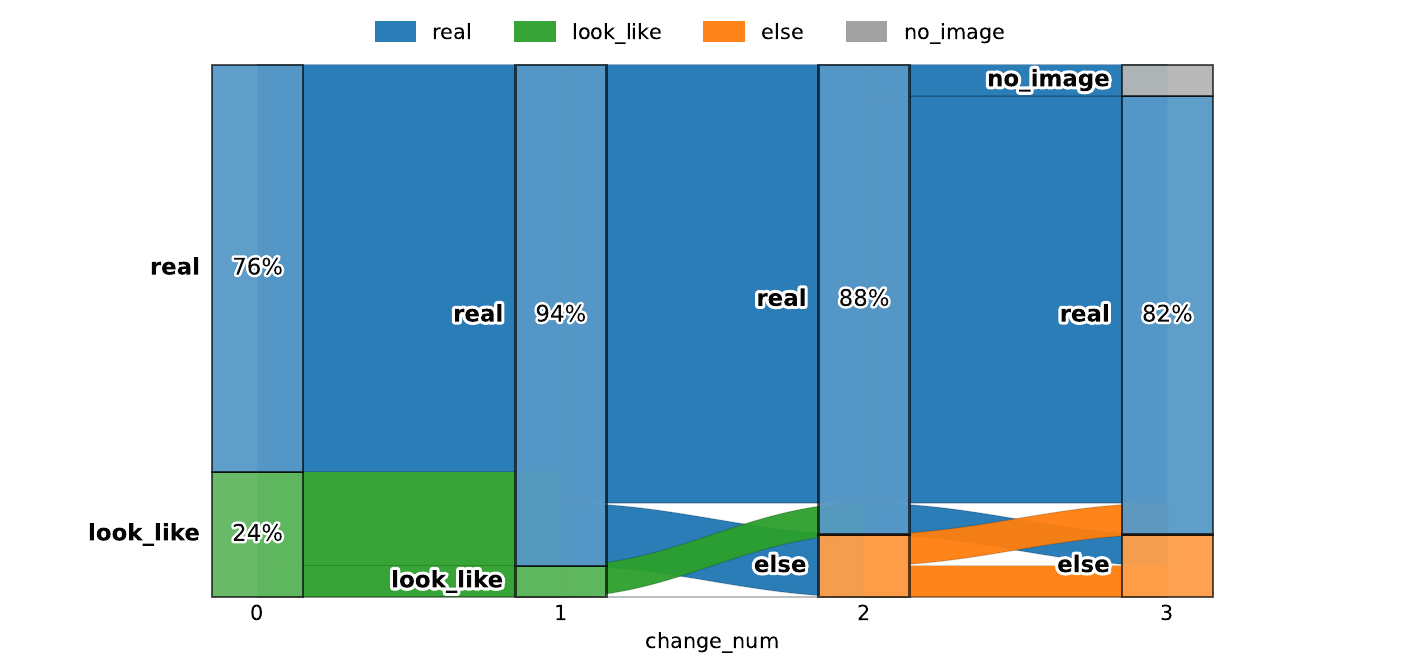"}
        \caption{AMT—real to more real.}
        \label{fig:AMT_real_more_real_new_changes}
    \end{minipage}

    \vspace{0.35cm}

    % Row 2
    \begin{minipage}{0.9\textwidth}
        \centering
        \includegraphics[width=\textwidth]{"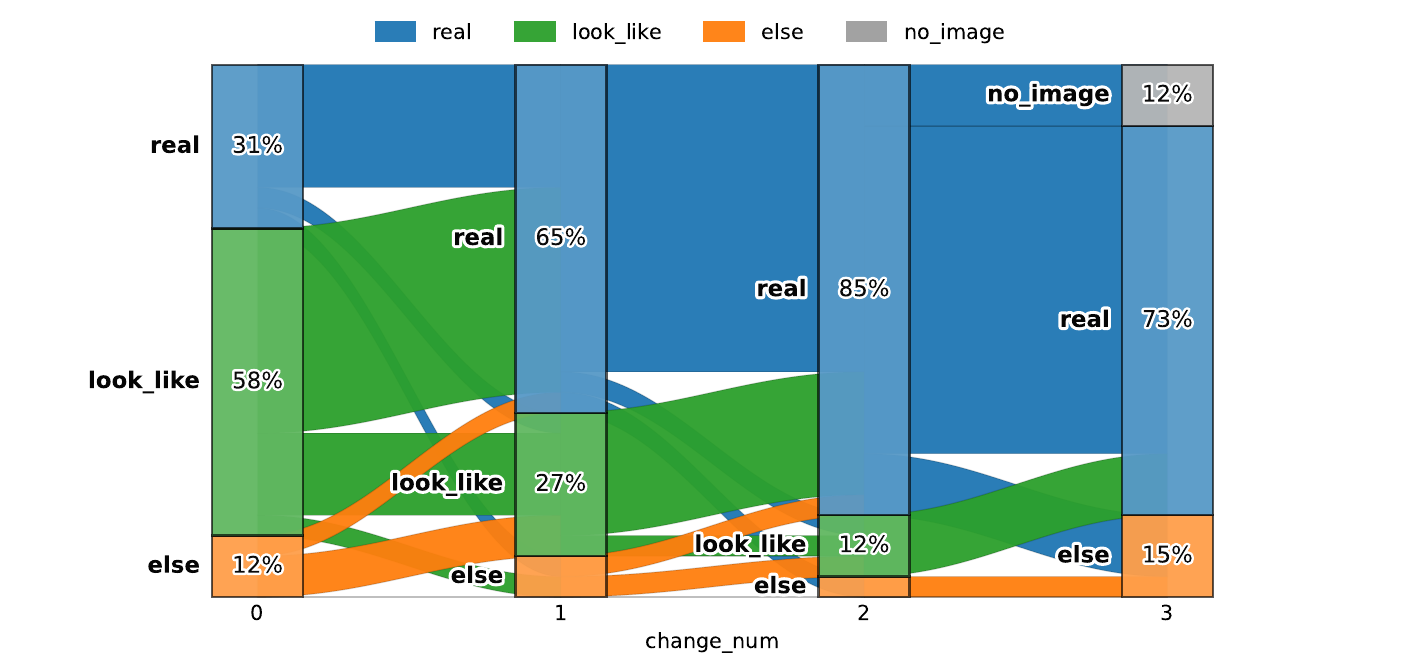"}
        \caption{AMT—Lookalike to real.}
        \label{fig:AMT_look_alike_to_real_changes}
    \end{minipage}
\end{figure}

In both plots (Fig.~\ref{fig:AMT_real_more_real_new_changes} and Fig.~\ref{fig:AMT_look_alike_to_real_changes}), The horizontal axis counts these changes. For example, if $\mathrm{NN}(0\le\alpha\le 0.05)$ remains constant, but switches at $\alpha=0.06$, then the first \emph{change} occurs at $\alpha=0.06$.\ef{move} \\
\emph{Real $\rightarrow$ more real} (Fig.~\ref{fig:AMT_real_more_real_new_changes}): each negative step reduces the share of images judged \textbf{lookalike} and
increases \textbf{real} and eventually \textbf{else}. After a single
change roughly three quarters of the initially \textbf{lookalike} items flip to
\textbf{real}; by two changes, \textbf{lookalike} labels are essentially gone, but also a small portion has left the class entirely and is marked
\textbf{else};
by three changes, another small portion has left the class entirely and is marked
\textbf{else}. This is the intended effect of pushing samples toward
lookalikes and, with larger shifts, beyond the class boundary. 
\emph{Lookalike $\rightarrow$ real} (Fig.~\ref{fig:AMT_look_alike_to_real_changes}):
moving in the negative direction has the mirror pattern. After one change,
about half of the \textbf{lookalike} items convert to \textbf{real}; after two,
most do; after three, nearly all remaining \textbf{lookalike} have become
\textbf{real}, with a small spillover to \textbf{else} at larger shifts, 
consistent with occasional overshoot past the class.

Together, these trends show that small shifts move items between
\textbf{real} and \textbf{lookalike} as designed, while large shifts
increasingly push items out of the class and elicit \textbf{else} judgments.

\noindent\textit{Note on \emph{no\_image}.} We sweep each prompt’s query embedding from
\(0\) up to a transition's threshold in steps of \(\alpha=0.01\) and then count how many
discrete “changes” the chain yields. Some prompts produce only two (or fewer) changes
before reaching the threshold. We select the threshold at the point where many embeddings have already moved out of the class (becoming “else”). For such prompts, the unused change slots (e.g., in
\(\text{changes}=3\)) do not have a corresponding image and are recorded as \emph{no\_image}.
This is a missing-item placeholder (not an AMT label) indicating that the sweep terminated before reaching that change index.

\subsection{Captioning}
We integrate the difference vector (Eq.~\eqref{eq:leave one out})
% \[
% \mathbf{d}^{(-k)}=\frac{1}{K-1}\sum_{i\ne k}\mathbf{d}_i,
% \]
into a CLIP prefix–captioner \cite{Mokady2021ClipCap} using a "leave one out" variant to avoid target leakage. Intuitively, moving the query by \(\pm k\Delta\) steers the caption semantics in a controlled way: subtracting \(k\Delta\) pulls captions toward \emph{real} descriptions of the class, while adding \(k\Delta\) pushes them toward \emph{lookalike} descriptions; larger \(k\) strengthens the effect.

\begin{figure}[H]
\centering

% Same visual height for all tiles, without exceeding width
\newlength{\tileH}
\setlength{\tileH}{0.28\textwidth} % ← tweak to match the previous (good) height

% Optional: tighten tables
\setlength{\tabcolsep}{4pt}
\renewcommand{\arraystretch}{1.05}

% ---- row 1 ----
\begin{subfigure}[t]{0.46\textwidth}
\centering
\begin{minipage}[t][\tileH][c]{\linewidth}
  \centering
  \includegraphics[height=\tileH,width=\linewidth,keepaspectratio]{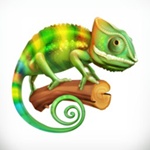}
\end{minipage}
\caption*{\textbf{A:} Image captioning—transitioning from lookalike to real }
\vspace{2pt}\footnotesize
\begin{tabular}{@{}r p{0.82\linewidth}@{}}
\multicolumn{1}{@{}c}{\scriptsize$\Delta$} 
% & \multicolumn{1}{@{}l}{\scriptsize$|\Delta|\downarrow$}
\\[-4pt]
$0$  & 3d rendering of a chameleon on a white background. \\
$-1$ & Biological species on a branch. \\
$-2$ & Biological species on a branch. \\
$-3$ & A close up of a green and tan chameleon. \\
$-4$ & What are the different types of amphibians?. \\
\end{tabular}
\end{subfigure}\hfill
\begin{subfigure}[t]{0.46\textwidth}
\centering
\begin{minipage}[t][\tileH][c]{\linewidth}
  \centering
  \includegraphics[height=\tileH,width=\linewidth,keepaspectratio]{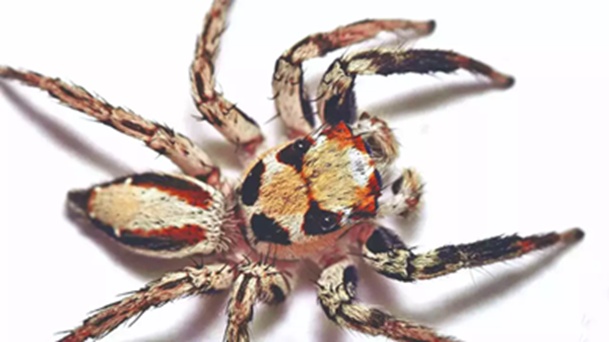}
\end{minipage}
\caption*{\textbf{B:} Image captioning—transitioning from real to lookalike}
\vspace{2pt}\footnotesize
\begin{tabular}{@{}r p{0.82\linewidth}@{}}
\multicolumn{1}{@{}c}{\scriptsize$\Delta$}
% & \multicolumn{1}{@{}l}{\scriptsize$|\Delta|\uparrow$}
\\[-4pt]
$0$  & A close-up of a spider's web. \\
$+1$ & A spider with a red face. \\
$+2$ & A spider made from a piece of paper. \\
$+3$ & This is a sculpture of a spider. \\
$+4$ & Handmade sculpture of a spider. \\
\end{tabular}
\end{subfigure}

\vspace{0.8em}

% ---- row 2 ----
\begin{subfigure}[t]{0.46\textwidth}
\centering
\begin{minipage}[t][\tileH][c]{\linewidth}
  \centering
  \includegraphics[height=\tileH,width=\linewidth,keepaspectratio]{images/image-captioning/image_captioning-spider-real_to_look_alike-just_image}
\end{minipage}
\caption*{\textbf{C:} Image captioning—transitioning from real to “more real.”}
\vspace{2pt}\footnotesize
\begin{tabular}{@{}r p{0.82\linewidth}@{}}
\multicolumn{1}{@{}c}{\scriptsize$\Delta$}
% & \multicolumn{1}{@{}l}{\scriptsize$|\Delta|\uparrow$}
\\[-4pt]
$0$  & A close-up of a spider's web. \\
$-1$ & Close up of a jumping spider. \\
$-2$ & A close-up of a jumping spider. \\
$-3$ & A close-up of a jumping spider. \\
$-4$ & A close up of a male animal. \\
\end{tabular}
\end{subfigure}\hfill
\begin{subfigure}[t]{0.46\textwidth}
\centering
\begin{minipage}[t][\tileH][c]{\linewidth}
  \centering
  \includegraphics[height=\tileH,width=\linewidth,keepaspectratio]{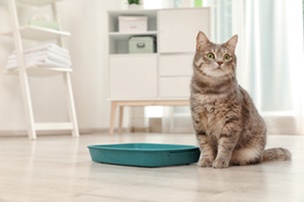}
\end{minipage}
\caption*{\textbf{D:} Image captioning—transitioning from real to lookalike}
\vspace{2pt}\footnotesize
\begin{tabular}{@{}r p{0.82\linewidth}@{}}
\multicolumn{1}{@{}c}{\scriptsize$\Delta$}
% & \multicolumn{1}{@{}l}{\scriptsize$|\Delta|\uparrow$}
\\[-4pt]
$0$  & Cat sitting on the floor and eating bowl. \\
$+1$ & Cat sitting on the floor and eating bowl. \\
$+2$ & Cat in a bowl — cute idea for a small table. \\
$+3$ & Cat in a pot — ceramic cat. \\
$+4$ & Cat in a pot — ceramic. \\
\end{tabular}
\end{subfigure}

\caption{Captioning outputs at different embedding transitions. Each tile starts with the original caption (\(\Delta=0\)) and proceeds with increasing magnitude \( |\Delta| \).}
\label{fig:captioning_grid_2x2}
\end{figure}

Figure~\ref{fig:captioning_grid_2x2} illustrates this behavior in four examples. In \emph{A}(top left), subtracting the vector three times (\(-3\Delta\)) turns an initially ambiguous output into a clean caption of a \emph{real chameleon} in a similar setting. In \emph{B}(top right), the baseline caption is wrong, yet adding \(+2\Delta\) or \(+3\Delta\) reliably produces captions of a \emph{lookalike spider}, as desired. Using the same source image in \emph{C}(bottom left) but reversing the direction, \(-\Delta\), \(-2\Delta\), or \(-3\Delta\) yields captions of a \emph{real spider} despite the original error. Finally, in \emph{D}(bottom right), starting from a \emph{real cat}, adding \(+3\Delta\) and \(+4\Delta\) shifts the captions toward a \emph{lookalike cat}.

Taken together, these transitions show smooth, monotonic control over caption semantics: small negative shifts favor \emph{real}, small positive shifts favor \emph{lookalike}, and increasing the magnitude \(k\) amplifies the intended change, even when the baseline caption is initially incorrect.

\section{Conclusion}

This study addresses an interesting aspect of computer vision: vision–language models such as CLIP~\cite{radford2021learning} often struggle to distinguish real objects from their lookalikes. 
We introduced the multi-category \textbf{RoLA} dataset and estimated a \emph{Real/Look-alike Direction} in CLIP’s embedding space by computing, for each category, the difference between the mean embeddings of look-alike and real images, and averaging these differences across categories to obtain a leave-one-out mean-difference vector. 
Applying this direction to both image and text embeddings improves discrimination in image retrieval on Conceptual~12M and produces consistent shifts in image captioning. 
Overall, we show that a simple, data-efficient direction learned from class-wise means enhances both retrieval and captioning performance and generalizes effectively across categories. %One direction for future work is specialization for visually similar categories (e.g., dog vs. wolf). This likely requires targeted data that captures fine-grained discriminative cues (attributes/parts) and training objectives that emphasize those cues (pairwise margins, hierarchical taxonomic priors, or attribute-aware prompts).

 % \JOVkeywords{vision-language models, CLIP, object recognition, prompt engineering, image–text retrieval, captioning}
 
% ---------------------- References ----------------------
\bibliography{references}   % <-- replace with your .bib file name
\bibliographystyle{ieeetr}

\section{Supplementary Material}

\subsection{Human Evaluation}\label{sec:Human Evaluation}

\begin{figure}[H]
\centering
\setlength{\hitH}{0.33\textwidth}

\begin{subfigure}[t]{0.48\textwidth}
  \centering
  \includegraphics[height=\hitH,keepaspectratio]%
    {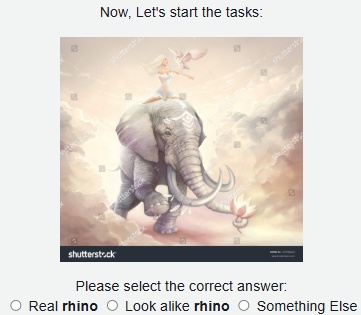}
  \subcaption{HIT example — rhino}
  \label{fig:hit-rhino}
\end{subfigure}\hfill
\begin{subfigure}[t]{0.48\textwidth}
  \centering
  \includegraphics[height=\hitH,keepaspectratio]%
    {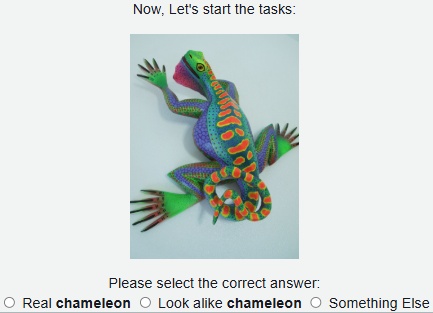}
  \subcaption{HIT example — chameleon}
  \label{fig:hit-chameleon}
\end{subfigure}

\caption{AMT HIT examples. \protect\subref{fig:hit-rhino}: example of \emph{Something else} where the query class is \emph{rhino} but the image shows an \emph{elephant}. \protect\subref{fig:hit-chameleon}: example of \emph{Look-alike chameleon}.}
\label{fig:hit-examples}
\end{figure}

We used Amazon Mechanical Turk (AMT) to test whether our “real vs.\ lookalike” outputs align with human judgments and to calibrate the shift \(\alpha\).
Participation was restricted to AMT Masters with \(>\!90\%\) approval. Each item showed an image and class name; workers chose one label: \emph{real [class]}, \emph{lookalike [class]}, or \emph{something else}. Every item received five independent judgments and we used the majority vote. Figure~\ref{fig:hit-examples} shows two example HITs.

\begin{figure}[H]
\centering
\setlength{\hitH}{0.31\textwidth} % slightly smaller height to match new width

\begin{subfigure}[t]{0.31\textwidth}
  \centering
  \includegraphics[height=\hitH,keepaspectratio]{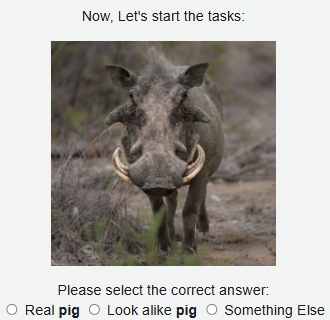}
  \subcaption{Warthog (a wild pig)}
  \label{fig:hit-pig-or-boar}
\end{subfigure}\hfill%
\begin{subfigure}[t]{0.31\textwidth}
  \centering
  \includegraphics[height=\hitH,keepaspectratio]{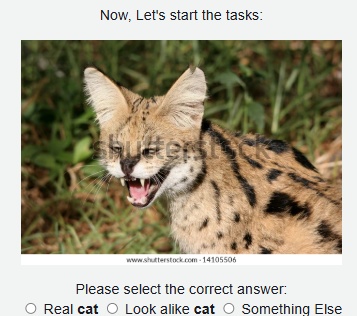}
  \subcaption{Serval (wild cat with big ears)}
  \label{fig:hit-cat-else1}
\end{subfigure}\hfill%
\begin{subfigure}[t]{0.31\textwidth}
  \centering
  \includegraphics[height=\hitH,keepaspectratio]{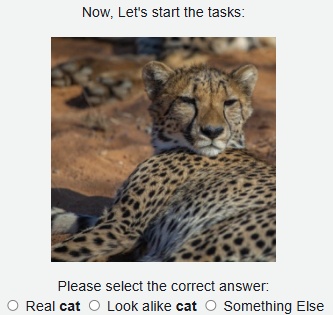}
  \subcaption{Cheetah}
  \label{fig:hit-cat-else2}
\end{subfigure}

\caption{AMT HIT disagreements caused by visually similar classes.
\protect\subref{fig:hit-pig-or-boar}: query \emph{pig} vs.\ image is a \emph{warthog};
\protect\subref{fig:hit-cat-else1}: query \emph{cat} vs.\ image is a \emph{serval};
\protect\subref{fig:hit-cat-else2}: query \emph{cat} vs.\ image is a \emph{cheetah}.}
\label{fig:hit-disagreements-examples}
\end{figure}

Across 500 items spanning 16 classes and multiple \(\alpha\) values, AMT judgments tracked our embedding transformation: small negative shifts tended to produce \emph{real} and small positive shifts \emph{lookalike}; very large shifts sometimes left the class and elicited \emph{something else}. Disagreements concentrated in visually similar categories (e.g., dog–wolf, cat–tiger; Fig.~\ref{fig:hit-disagreements-examples}). Class-specific directions worked best within their own class, while a small leave-one-out direction generalized well with only a modest drop in agreement.

\end{document}